\documentclass[runningheads]{llncs}
\usepackage[T1]{fontenc}
\usepackage{graphicx,verbatim}
\usepackage{amsmath}
\usepackage{cite}
\usepackage{booktabs}
\usepackage{orcidlink}
\usepackage{hyperref}
\usepackage{marvosym}
\usepackage{multirow}
\usepackage[table,xcdraw]{xcolor}
\usepackage{amssymb}
\begin{document}
\title{HybridMamba: A Dual-domain Mamba for 3D Medical Image Segmentation}

\author{Weitong Wu\inst{1} %index{Wu, Weitong}
\and
Zhaohu Xing\inst{1} %index{Xing, Zhaohu}
\and
Jing Gong\inst{2,3} %index{Gong, Jing}
\and
Qin Peng\inst{2,3} %index{Peng, Qin}
\and
Lei Zhu\inst{1,4} $^\textrm{\Letter}$ \\ %index{Zhu, Lei}
}  %% Added for anonymized MICCAI 2025 submission
\authorrunning{{Weitong Wu, Zhaohu Xing, Jing Gong, Qin Peng, and Lei Zhu $^\textrm{\Letter}$}}
 
\institute{The Hong Kong University of Science and Technology (Guangzhou), \\ Guangzhou, China \\
\and
Department of Radiology, Fudan University Shanghai Cancer Center, \\ Shanghai, China
\and
Department of Oncology, Shanghai Medical College, Fudan University, \\ Shanghai, China
\and
The Hong Kong University of Science and Technology, Hong Kong, China \\
\email{leizhu@ust.hk} 
}

\maketitle              % typeset the header of the contribution
\begin{abstract}
In the domain of 3D biomedical image segmentation, Mamba exhibits the superior performance for it addresses the limitations in modeling long-range dependencies inherent to CNNs and mitigates the abundant computational overhead associated with Transformer-based frameworks when processing high-resolution medical volumes. However, attaching undue importance to global context modeling may inadvertently compromise critical local structural information, thus leading to boundary ambiguity and regional distortion in segmentation outputs. Therefore, we propose the HybridMamba, an architecture employing dual complementary mechanisms: 1) a feature scanning strategy that progressively integrates representations both axial-traversal and local-adaptive pathways to harmonize the relationship between local and global representations, and 2) a gated module combining spatial-frequency analysis for comprehensive contextual modeling. Besides, we collect a multi-center CT dataset related to lung cancer. Experiments on MRI and CT datasets demonstrate that HybridMamba significantly outperforms the state-of-the-art methods in 3D medical image segmentation.

% Authors must provide keywords and are not allowed to remove this Keyword section.

\end{abstract}
\section{Introduction}
In clinical diagnostic workflows, achieving voxel-level precision in pathological region segmentation from 3D medical imaging modalities (e.g., CT, MRI) constitutes a critical prerequisite for quantitative disease characterization~\cite{xing2025diff}. While deep learning architectures have demonstrated remarkable potential in medical image segmentation, thereby reducing inter-observer variability and clinician workload, there still exist fundamental limitations. Traditional Convolutional Neural Networks (CNNs) suffer from structural constraints in modeling long-range dependencies within medical image segmentation due to their intrinsic locality~\cite{chen2021transunet}. Deep learning architectures have shown great promise in medical image segmentation by minimizing inter-observer variability and clinician workload. However, traditional Convolutional Neural Networks (CNNs) are fundamentally limited by structural constraints that hinder their ability to model long-range dependencies due to their inherent locality~\cite{chen2021transunet}. Recent attempts like 3D UX-Net~\cite{lee20223d} endeavor to mitigate this issue by expanding effective receptive fields through large-kernel convolutions, but still have constraints in modeling global relationships. 

The emergence of Transformer architecture~\cite{vaswani2017attention} has notably enhanced the modeling of global context through self-attention mechanisms, effectively addressing the limitations of CNNs in capturing long-range dependencies. For instance, UNETR~\cite{hatamizadeh2022unetr} merges the Vision Transformer (ViT)~\cite{dosovitskiy2020image} into the encoder to capture contextual information and utilizes a convolutional decoder with multi-scale skip connections to learn local features together for 3D medical image segmentation.
Similarly, SwinUNETR~\cite{hatamizadeh2021swin} employs the SwinTransformer~\cite{liu2021swin} in its encoder to enable efficient extraction of features at multiple resolutions. Nonetheless, these Transformer-based methods imposes significant scalability constraints for high-resolution 3D medical imaging analysis. This computational bottleneck manifests particularly in memory-intensive segmentation tasks like biomedical image segmentation,resulting in suboptimal throughput rates during both training and inference phases despite recent hardware advancements.

Emerging from State Space Model, Mamba~\cite{gu2023mamba,wang2024serp,wu2024rainmamba} presents an innovative approach to long-range dependency learning through selective mechanism and hardware-efficient algorithms. This paradigm achieves CNN-level training stability coupled with RNN-like inference efficiency~\cite{gu2020hippo}, all maintained within linear computational complexity. Current advances demonstrate growing applications of Mambas in medical imaging analysis: In 2D contexts, U-Mamba~\cite{ma2024u} augments standard nnUNet frameworks~\cite{isensee2021nnu} by integrating directional scanning modules into encoder pathways, significantly enhancing the performance in medical image segmentation. Moreover, Swin-UMamba~\cite{liu2024swin} strategically incorporates large-scale ImageNet pretrained representations, thereby enhanceing Mamba's efficacy in clinical image segmentation scenarios. For 3D medical image analysis, SegMamba~\cite{xing2024segmamba} pioneers voxel-level contextual modeling through its tri-orientated Mamba (ToM) module for feature modeling and gated spatial convolution (GSC) module for the representation enhancement of spatial features. However, Existing 2D methodologies fail to address cross-slice correlations inherent in volumetric datasets effectively; meanwhile SegMamba exhibit suboptimal coordination between localized feature retention and global context integration when processing hierarchically partitioned image patches.

In this paper, we propose HybridMamba, a hierarchical architecture addressing two fundamental challenges in 3D medical image analysis: 1) multi-resolution contextual preservation across spatial scales; and 2) synergistic integration of frequency-spatial feature representations. To our best knowledge, this is the first method to deploy the feature extraction strategy in both the frequency domain and the spatial domain to facilitate more accurate and robust representations on multiple 3D medical image segmentation tasks. Consider the non-negligibility of relationships between global and local information, we design a fused ergodic mechanism named SoMamba (Slice-oriented Mamba) and LoMamba(Local-oriented Mamba) to increase the sequential modeling of 3D features. Following this, we further propose the gated mechanism that dynamically weights frequency-transformed features against spatially encoded patterns prior to cascaded Mamba processing stages. We verify the superior performance of the proposed HybridMamba on both public MRI dataset and our collected CT dataset. The results of experiments showcase the efficiency of our method.

\section{Method}
\begin{figure}[t]
\includegraphics[width=0.95\textwidth]{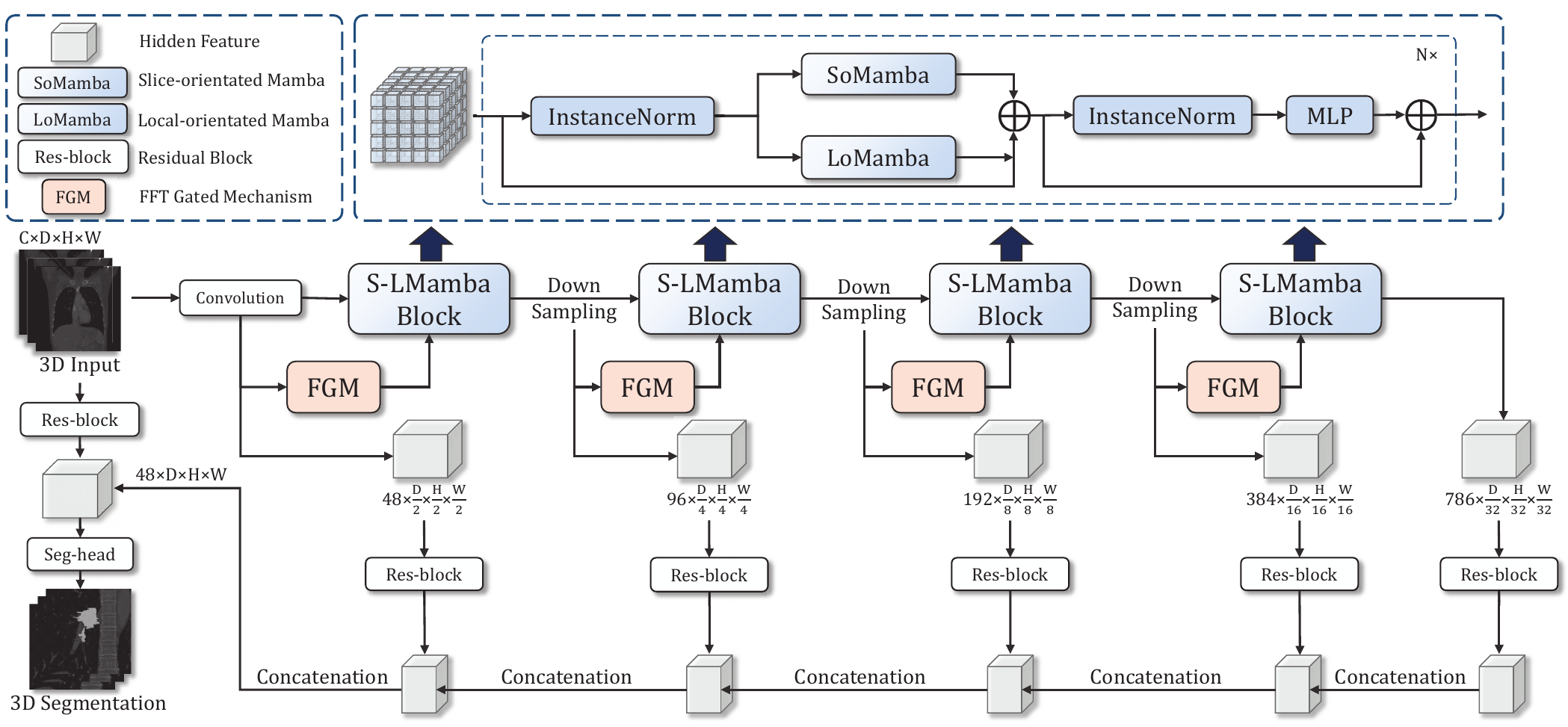}
\centering
\caption{The overview of the proposed HybridMamba. The encoder consists of a multiple S-LMamba blocks for balancing local and global features and an FFT Gate Mechanism (FGM) for merging the features from spatial and frequency domain dynamically according to the characteristics of different layers.} 
\label{overview}
\end{figure}

In this section, we introduce the implementation approach of HybridMamba, which is built upon SegMamba~\cite{xing2024segmamba} with the particular improvements on encoder. We incorporate the frequency domain feature into encoder modeling by utilizing the gated mechanism before transforming from two kinds of traversing pathways to sequence. Fig.~\ref{overview} demonstrates the overview of HybridMamba. The details of the encoder will be further described in this section. 

\subsection{Slice-Local Mamba (S-LMamba) Block}
It is of great necessity to utilize Mamba to model long-range dependencies within 3D high-resolution biomedical image segmentation~\cite{xing2023diff,xing2024hybrid,wang2024dual}. While transformer architectures~\cite{wu2024semi,wang2024video} effectively capture global information, they impose substantial computational costs when processing excessively long feature sequences. Accordingly, Mamba is utilized especially for high-resolution biomedical image field. It fully leverages long-distance modeling while being exponentially more computationally efficient than the Transformer framework. However, prioritizing long-distance dependencies overlooks the extraction of local segmentation information within and between slices, potentially leading to incoherent partitioning and inadequate semantic understanding in the model. Drawing inspiration from the scanning method presented in Local Mamba~\cite{huang2024localmamba}, we design an S-LMamba Block to partition all slices of the medical image, generating windows of a desired size according to different scales in each layer. We also recognize the importance of long-distance dependencies throughout the training process. Therefore, we introduce a slice-oriented traversing strategy to enhance the learning of contextual information within the medical image. This approach allows Mamba to extract more relevant segmentation information from the sequences.

\begin{figure}[t]
\centering
\includegraphics[width=0.95\textwidth]{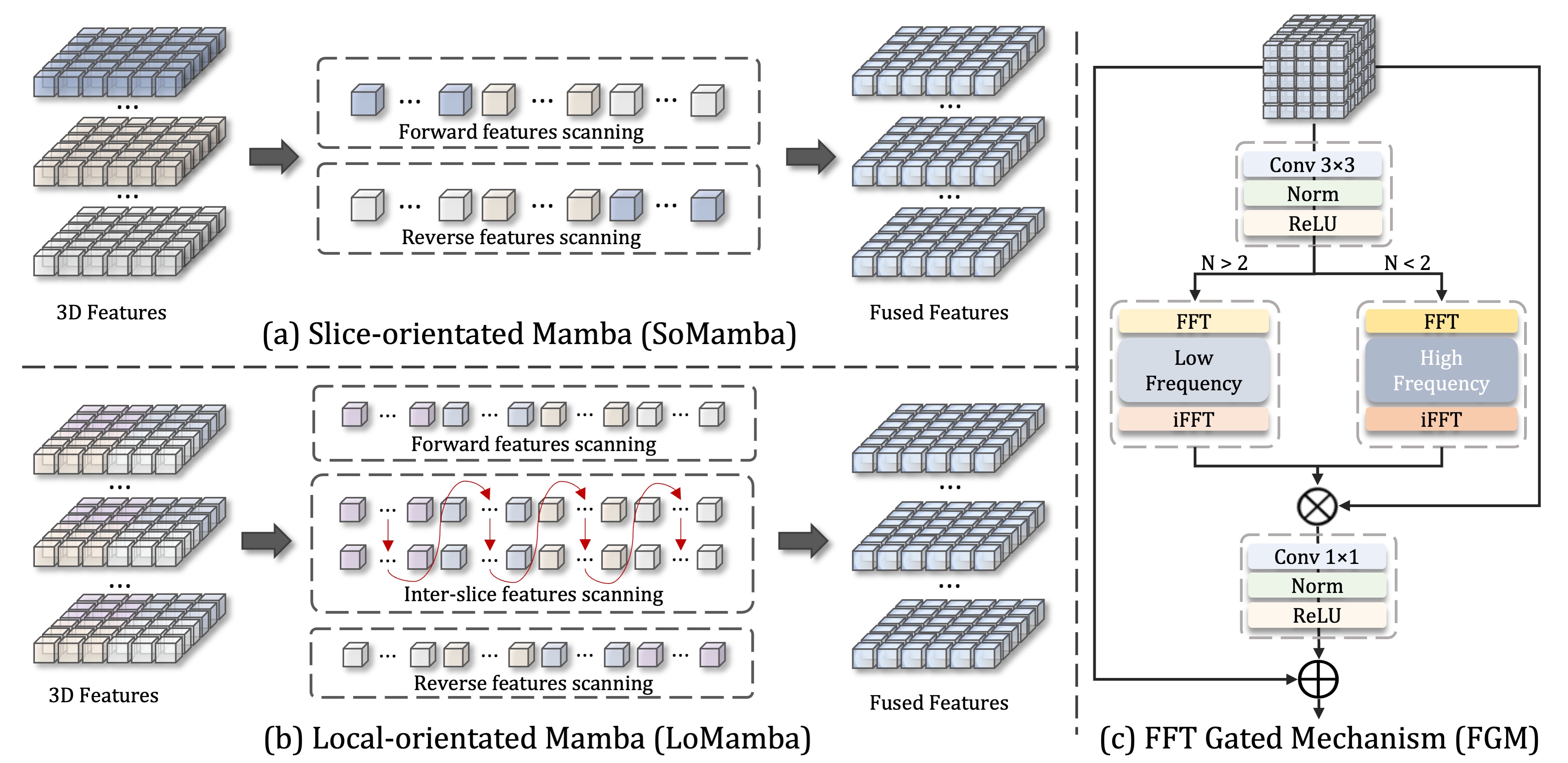}
\caption{The left side (a) and (b) shows the detailed implementation in SoMamba and LoMamaba. The right side (c) demonstrates the specific layers in the FGM.} 
\label{solofgm}
\end{figure}

The encoder illustrated in Fig.~\ref{overview} is composed of multiple S-LMamba blocks and FGM modules. After the initial preprocessing of the input 3D volume $I\in\mathbb{R}^{48\times{D}\times{H}\times{W}}$ busing a large convolutional layer with a kernel size of ${7}\times{7}\times{7}$, the resulting 3D feature $x_{0}\in\mathbb{R}^{48\times\frac{D}{2}\times\frac{H}{2}\times\frac{W}{2}}$ is processed through the S-LMamba blocks and FGM modules, along with subsequent down-sampling layers. For the $n^{th}$ S-LMamba Block, the computational operation can be defined as:
\begin{equation}
\quad\tilde{x}_{n}^{l}=\text{MLP}(\text{IN}(\text{SoMamba}\left(\mathrm{LN}\left(\hat{x}_{n}^{l}\right)\right)+\text{LoMamba}\left(\mathrm{LN}\left(\hat{x}_{n}^{l}\right)\right)+\hat{x}_{n}^{l}))+\tilde{x}_{n}^{l},  
\end{equation}
where the SoMamba and LoMamba refer to the proposed Slice-oriented and Local-oriented Mamba module, respectively, which will be discussed next. ${l}\in\left\{0,1,...,n-1\right\}$, LN refers to layer normalization, and IN refers to instance normalization. MLP refers to the multi-layer perceptron for enriching the feature representation.

\noindent
\textbf{Slice-orientated Mamba (SoMamba) and Local-orientated Mamba (LoMamba).} 
The Mamba layer captures feature dependencies by flattening the 3D features into a 1D sequence. Managing the order of this flattening process is crucial, as it directly impacts the model's learning efficiency. To optimize the arrangement of the flattened 1D sequence, we introduce the Slice-orientated Mamba (SoMamba) and Local-orientated Mamba (LoMamba) module illustrated in part (a) and (b) of Fig.~\ref{solofgm}. 
\begin{equation}
\text{SoMamba}(x)=\text{Mamba}(x_f)+\text{Mamba}(x_r), \label{so}
\end{equation}
\begin{equation}
\text{LoMamba}(x)=\text{Mamba}(x_{lf})+\text{Mamba}(x_{lr})+\text{Mamba}(x_{ls}),\label{lo}
\end{equation}
where Mamba denotes the Mamba layer to model the global information within the sequence. ${f}$ and ${r}$ in Eq.~\ref{so} denote forward and reverse direction respectively. ${lf}$, ${lr}$, and ${ls}$ in Eq.\ref{lo} refer to local-window forward direction, local-window reverse direction, and local-window across slices, correspondingly.

\begin{figure}[b]
\centering
\includegraphics[width=0.95\textwidth]{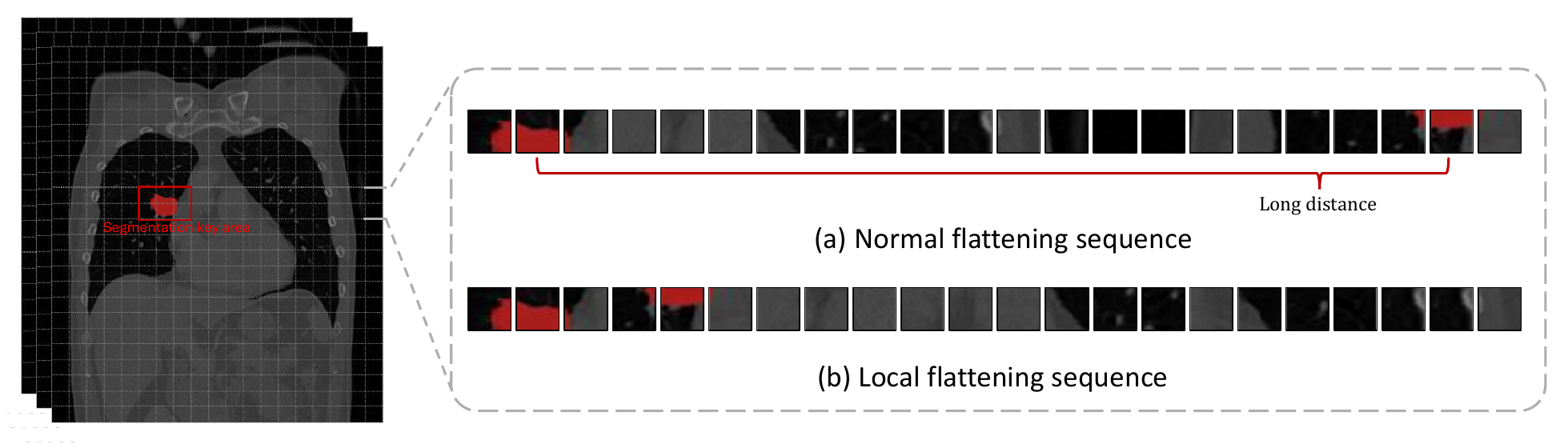}
\caption{Take $3\times3$ local window size around segmentation key area on one slice as an example. Sequence (a) showcases the normal flatten method, which traverse the whole image patches from beginning to end. Sequence (b) demonstrates the local flatten method. } \label{scan}
\end{figure}

For SoMamba, we design a traversing path that spans the entire slice in both forward and reverse directions to compute feature dependencies, aiming to effectively capture the global information inherent in high-dimensional features. Additionally, we observe the presence of short-distance dependencies within the overall medical image, particularly as lesion areas may constitute a smaller proportion of the total. To address this, we design a Local modeling strategy known as LoMamba to focus on the adjacent pixels with the same semantic region, which empower the model to better aggregate the features because of more compact physical position of key information after flattening into a 1D sequence. 

Fig.~\ref{scan} showcases the distinction between normal scanning and local scanning modes using the example of the local flattening sequence with a window size of three. In practice, the size of the local window is determined dynamically by the proportion and location of key information areas within the overall medical image. To be specific, the sequence through the standard flattening method exhibits a considerable distance between neighboring pixels that contain important segmentation information. In contrast, the continuous key segmentation pixels flattened by local windows of size three manifesting a strong distance coherence to a certain extent.

\subsection{FFT Gated Mechanism (FGM)}
Fast Fourier Transform (FFT)~\cite{bergland1969guided} is leveraged to calculate the frequency values from spatial domain. In cases of CT images with poor contrast and high noise, as well as MRI images affected by artifacts, frequency information from high level and low level can correspondingly provide some boundary and shape cues to the model~\cite{zhou2023xnet}, which enables it to capture a wider range of feature representations, thus increasing the robustness. Furthermore, as highlighted in~\cite{xu2019frequency}, the deeper layers of deep neural network tend to retain more low-frequency information. To capitalize on this, we design the FFT Gated Mechanism (GSM) to integrate the frequency and spatial features to enhance the performance of Mamba model. As illustrated in subgraph (c) of Fig.~\ref{solofgm}, the input 3D features are first fed into a convolution block, which contains a convolution, a normalization, and an activation layer. Then, the feature are transformed into the Fourier domain according to different layers, being extracted high and low frequency leveraging learnable thresholds within a filter. Subsequently, conducting inverse FFT to transform the required frequency feature back. Finally, a convolution block is used to further fuse the frequency features and spatial features after gated multiplication, while a residual connection is utilized to reuse the input features.

\begin{equation}
{x}_\text{fre}=\text{IFFT}(\,\text{Filter}(\,\text{FFT}\,(x))), \quad {x_\text{out}} = {x_{s}} * \text{gate} + {x_\text{fre}} * (1-\text{gate}),
\end{equation}
\begin{equation}
\left.\text{Filter}=
\begin{cases}
 x*(|x|<f_\text{low}),x\in \text{low-level\,frequency}, \\
 x*(|x|>f_\text{high}),x\in \text{high-level\,frequency},
\end{cases}\right.
\end{equation}
where $\text{gate}=\text{Conv}_{3\times3\times3}({x}_\text{fft}, {x}_{s})$ is used to get the prior embedding to realize the coordination of features in spatial domain and frequency domain. $f_{low}$, $f_{high}$ denote the learnable thresholds of the filter with the initial values set to 0.1 and 0.9 respectively.

\section{Experiments}

\subsection{Datasets and Implementation Details}
\noindent
\textbf{BraTS2023 Dataset~\cite{kazerooni2024brain, bakas2017advancing, menze2014multimodal}.} This dataset comprises 1,251 3D MRI volumes of the brain. Each volume is presented in four different imaging modalities: T1, T1Gd, T2, and T2-FLAIR. Three segmentation targets are identified within each volume: Whole Tumor (WT), Enhancing Tumor (ET), and Tumor Core (TC).

\begin{figure}[t]
\centering
\includegraphics[width=0.95\textwidth]{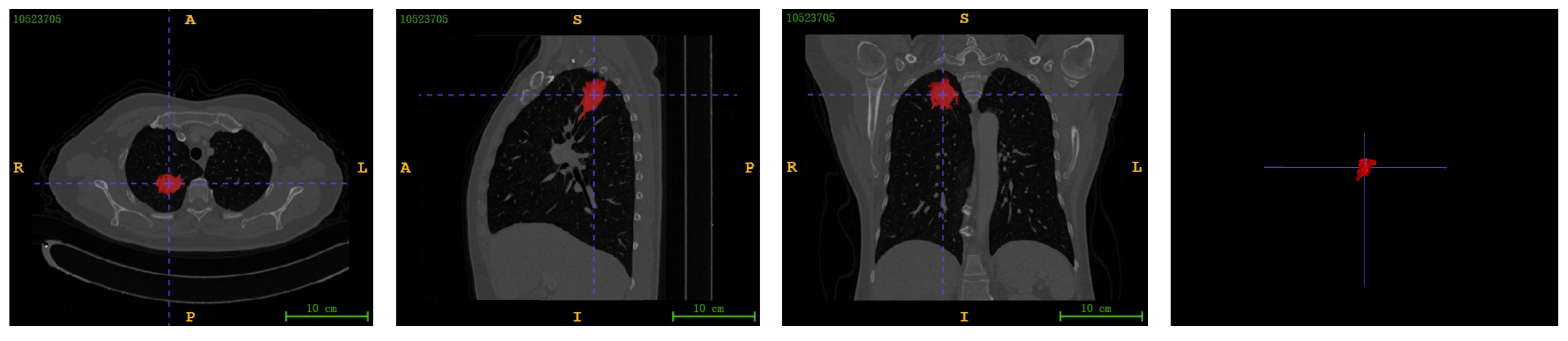}
\caption{The data visualization for LC dataset, which highlights the challenges posed by small lesions, which can be difficult to detect and analyze. } \label{dataset}
\end{figure}

\noindent
\textbf{Lung Cancer (LC) Dataset.} 
We gathered a collection of 828 3D chest CT scans from multiple centers, focusing on cases of central type lung carcinoma with small lesions. These cases pose inherent challenges in detection and analysis, as they appear as subtle anomalies within the imaging data. Each volume contains a single segmentation target, with the dominant lesion accurately annotated for each case. The visualization of this dataset is depicted in Fig~\ref{dataset}.

\noindent
\textbf{Evaluation and Metrics.}
In line with established approaches~\cite{xing2024segmamba}, we employ the Dice score (Dice) and Hausdorff Distance (HD95) to quantitatively assess our network's performance and compare it with state-of-the-art (SOTA) methods.

\noindent
\textbf{Implementation Details.}
Our model is developed using PyTorch 2.0.1 with CUDA 11.8 and Monai 1.3.0. During training, we apply a random crop size of 128 × 128 × 128 and utilize a batch size of 2 per GPU for each dataset. We employ cross-entropy loss for all experiments, using an SGD optimizer with a polynomial learning rate scheduler (initial learning rate set at 1e-4 and a decay rate of 3e-5). Each dataset undergoes 1000 training epochs, and we incorporate the following data augmentations: additive brightness, gamma correction, rotation, scaling, mirroring, and elastic deformation. All experiments are conducted on a cloud computing platform equipped with four NVIDIA GeForce RTX 4090 GPUs. For each dataset, we randomly assign 70\% of the 3D volumes for training, 10\% for validation, and the remaining 20\% for testing.

\begin{table}[b]
\caption{Quantitative comparison on BraTS2023 dataset. The bold value denotes the best performance.}\label{tab_com}
\centering
\resizebox{0.85\textwidth}{!}{
\begin{tabular}{ccccccccc|cc}
\toprule
\multicolumn{9}{c|}{BraTS2023}                                                                                                                               & \multicolumn{2}{c}{LC}          \\ 
\midrule
\multirow{2}{*}{Methods} & \multicolumn{2}{c}{WT}         & \multicolumn{2}{c}{TC}         & \multicolumn{2}{c}{ET}         & \multicolumn{2}{c|}{Avg}       & \multicolumn{2}{c}{Lung Cancer} \\& Dice↑          & HD95↓         & Dice↑          & HD95↓         & Dice↑          & HD95↓         & Dice↑          & HD95↓         & Dice↑          & HD95↓          \\ \midrule
SegresNet~\cite{myronenko20193d}                
& 92.02         & 4.07          & 89.10          & 4.08          
& 83.66         & 3.88          & 88.26          & 4.01         
& 71.56         & 55.03          \\
UX-Net~\cite{lee20223d}                   
& 93.13          & 4.56          & 90.03          & 5.68          
& 85.91          & 4.19          & 89.69          & 4.81          
& 72.63          & 59.06         \\
MedNeXt~\cite{roy2023mednext}                  
& 92.41          & 4.98         & 87.75          & 4.67          
& 83.96          & 4.51         & 88.04          & 4.72          
& 57.76          & 111.47        \\ \midrule
UNETR~\cite{hatamizadeh2022unetr}                    
& 92.19        & 6.17          & 86.39          & 5.29          
& 84.48        & 5.03          & 87.68          & 5.49          
& 65.12        & 101.7          \\
SwinUNETR~\cite{hatamizadeh2021swin}                
& 92.71          & 5.22          & 87.79          & 4.42          
& 84.21          & 4.48          & 88.23          & 4.70          
& 65.83          & 89.27          \\ \midrule
SegMamba~\cite{xing2024segmamba}                 
& 93.61          & \textbf{3.37} & 92.65          & 3.85          
& 87.71          & 3.48          & 91.32          & 3.56          
& 72.36          & 60.33          \\ \midrule
\rowcolor[HTML]{EFEFEF} 
Ours                     
& \textbf{94.10} & 3.78        & \textbf{92.84} & \textbf{3.30} & \textbf{88.83} & \textbf{3.35} & \textbf{91.92} & \textbf{3.48} & \textbf{75.34} & \textbf{44.52} \\ 
\bottomrule
\end{tabular}
}
\end{table}

\begin{figure}[h]
\centering
\includegraphics[width=0.95\textwidth]{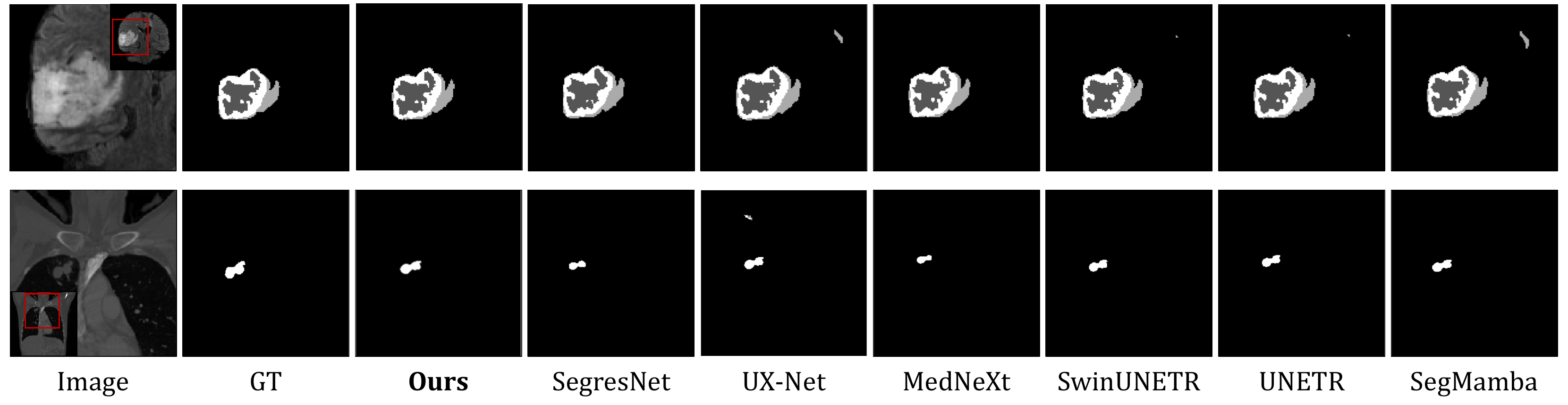}
\caption{Visual comparisons of proposed HybridMamba and state-of-the-art methods.} \label{visual}
\end{figure}

\subsection{Experimental Results}
\textbf{Compared Methods.} We evaluate our network by comparing it to six state-of-the-art (SOTA) 3D image segmentation methods, which includes three CNN-based approaches (SegResNet~\cite{myronenko20193d}, UX-Net~\cite{lee20223d}, MedNeXt~\cite{roy2023mednext}), two transformer-based methods (UNETR~\cite{hatamizadeh2022unetr} and SwinUNETR~\cite{hatamizadeh2021swin}), and one Mamba-based technique (SegMamba~\cite{xing2024segmamba}).

\noindent
\textbf{Quantitative Comparisons.} Table ~\ref{tab_com} summarizes the Dice score and HD95 for each modalities on BraTS2023 and LC dataset and the total average scores. For BraTS2023 dataset, SegMamba, the Mamba-based method, achieves the best performance among the comparison methods, with an average Dice of 91.32\% and an average HD95 of 3.56. In comparison, our HybridMamba achieves the highest Dice of 94.10\%, 92.84\%, and 88.83\% on WT, TC, and ET, respectively, and the best HD95 with 3.30 on TC and 3.35 on ET except for 3.78 on WT. All in all, the total average scores shows the best segmentation robustness. In addition, our HybridMamba outperforms the SOTA method exceeding 2.98\% for Dice and 15.81 for HD95 on LC dataset, getting 75.34\% and 44.52\% respectively. This proves the most effectiveness of HybridMamba compared to other approaches.

\noindent
\textbf{Visual Comparisons.} We choose six comparative methods for visual assessment on two datasets to evaluate image segmentation performance. As shown in Fig.~\ref{visual}, our HybridMamba effectively delineates the boundary of each tumor region in the BraTS2023 dataset. Similarly, our approach successfully identifies cancerous areas in the LC dataset. The segmentation results demonstrate enhanced consistency compared to other state-of-the-art techniques.

\noindent
\textbf{Ablation Studies.}
Table~\ref{tab_ab} confirms the effectiveness of both S-LMamba (M1) and FGM (M2) modules on the LC dataset. S-LMamba improves SegMamba's Dice by 1.47\% to 73.83\% and reduces HD95 by 9.45 to 69.78. FGM further boosts Dice to 74.98\% and reduces HD95 by 18.35 to 51.43. HybridMamba, combining both modules, achieves optimal results with 75.34\% Dice and 44.52 HD95.

\begin{table}[h]
\caption{Ablation study for different modules on LC dataset.}\label{tab_ab}
\centering
\resizebox{0.5\textwidth}{!}
{\begin{tabular}{ccccc}
\hline
& \multicolumn{2}{c}{Modules} 
&                          &                         \\
\multirow{-2}{*}{Methods} 
& S-LMamba       & FGM       
& \multirow{-2}{*}{Dice ↑} & \multirow{-2}{*}{HD95↓} \\     \hline
SegMamba                  &                 &           
& 72.36                   & 69.78                   \\
M1                        & \checkmark               &           
& 73.83                   & 60.33                   \\
M2                        &                 & \checkmark        
& 74.98                   & 51.43                   \\     \hline
\rowcolor[HTML]{EFEFEF} 
Ours                      & \checkmark               & \checkmark      & \textbf{75.34}          & \textbf{44.52}     \\    \hline
\end{tabular}
}
\end{table}

\section{Conclusion}
In this work, we have developed HybridMamba to enhance the 3D biomedical segmentation task.
Specifically, our network makes two primary contributions. First, we devise the S-LMamba block to effectively balance the modeling of global and local dependencies. Second, we aggregate frequency features with spatial features to enhance the representation of the model. Experimental results on two datasets demonstrate that our framework clearly outperforms SOTA methods in terms of 3D medical image segmentation task.

\subsubsection{Acknowledgments.}
This research is supported by the Guangdong Science and Technology Department (No. 2024ZDZX2004), the Guangdong Provincial Key Lab of Integrated Communication, Sensing and Computation for Ubiquitous Internet of Things(No.2023B1212010007), the National Natural Science Foundation of China (No. 82001903), the Natural Science Foundation of Shanghai (No. 25ZR1402077 and 21ZR1414200), and the Excellent Young Talents Project of Shanghai Public Health Three-year (2023-2025) Action Plan (No. GWVI-11.2-YQ48).

\subsubsection{Disclosure of Interests.}
The authors declare that they have no competing interests.

\bibliographystyle{splncs04}
\bibliography{ref}




% LNCS Homepage, \url{http://www.springer.com/lncs}, last accessed 2023/10/25
% \end{thebibliography}
\end{document}